%% file: iclr2025_conference.tex
\documentclass{article}
\usepackage{iclr2025_conference,times}

\input{math_commands.tex}

\usepackage{hyperref}
\usepackage{url}
\usepackage{booktabs}
\usepackage{wrapfig}
\usepackage{caption}
\usepackage{multirow}
\usepackage{graphicx}
\usepackage{subcaption}
\usepackage{graphicx}
\usepackage{makecell}
\usepackage{pgfplots}
\usepackage{xcolor}
\usepackage{fontawesome5}
\usepackage{lipsum}
\pgfplotsset{compat=1.3}
\usepackage{booktabs}
\usepackage{adjustbox}
\newcommand{\tabincell}[2]{\begin{tabular}{@{}#1@{}}#2\end{tabular}}
\usepackage{xcolor}

\definecolor{deptA}{HTML}{6A1E1E}
\definecolor{deptB}{HTML}{1F5E3B}
\definecolor{deptC}{HTML}{1F2A44}

\newcommand{\name}{Sigma-MoE-Tiny}

\title{\name{} Technical Report}

\author{\textsc{Sigma} v-Team\\Microsoft Research}

\iclrfinalcopy
\begin{document}

\maketitle

\input{section/0_abs}

\input{section/1_intro}

\input{section/2_arch}
\input{section/3_pre}
\input{section/4_post}
\input{section/5_con}

\bibliography{iclr2025_conference}
\bibliographystyle{iclr2025_conference}

\appendix
\newpage
\input{section/appendix}

\end{document}

%% file: math_commands.tex
\usepackage{amsmath,amsfonts,bm}

\def\eqref#1{equation~\ref{#1}}

\def\1{\bm{1}}

\DeclareMathAlphabet{\mathsfit}{\encodingdefault}{\sfdefault}{m}{sl}
\SetMathAlphabet{\mathsfit}{bold}{\encodingdefault}{\sfdefault}{bx}{n}



%% file: section/0_abs.tex
\begin{abstract}
Mixture-of-Experts (MoE) has emerged as a promising paradigm for foundation models due to its efficient and powerful scalability. In this work, we present \name{}, an MoE language model that achieves the highest sparsity compared to existing open-source models.
\name{} employs fine-grained expert segmentation with up to 96 experts per layer, while activating only one expert for each token, resulting in 20B total parameters with just 0.5B activated. The major challenge introduced by such extreme sparsity lies in expert load balancing. We find that the widely-used load balancing loss tends to become ineffective in the lower layers under this setting. To address this issue, we propose a progressive sparsification schedule aiming to balance expert utilization and training stability.
\name{} is pre-trained on a diverse and high-quality corpus, followed by post-training to further unlock its capabilities. The entire training process remains remarkably stable, with no occurrence of irrecoverable loss spikes.
Comprehensive evaluations reveal that, despite activating only 0.5B parameters, \name{} achieves top-tier performance among counterparts of comparable or significantly larger scale.
In addition, we provide an in-depth discussion of load balancing in highly sparse MoE models, offering insights for advancing sparsity in future MoE architectures.
\end{abstract}

\begin{center}
    \begin{tabular}{cl}
    \faGithub & \url{https://github.com/microsoft/ltp-megatron-lm} \\
     \faGlobe & \url{https://qghuxmu.github.io/Sigma-MoE-Tiny}

    \end{tabular}
\end{center}

\vspace{10pt}
\begin{figure*}[h]
    \hspace{-5pt}
    \centering
    \includegraphics[width=\textwidth]{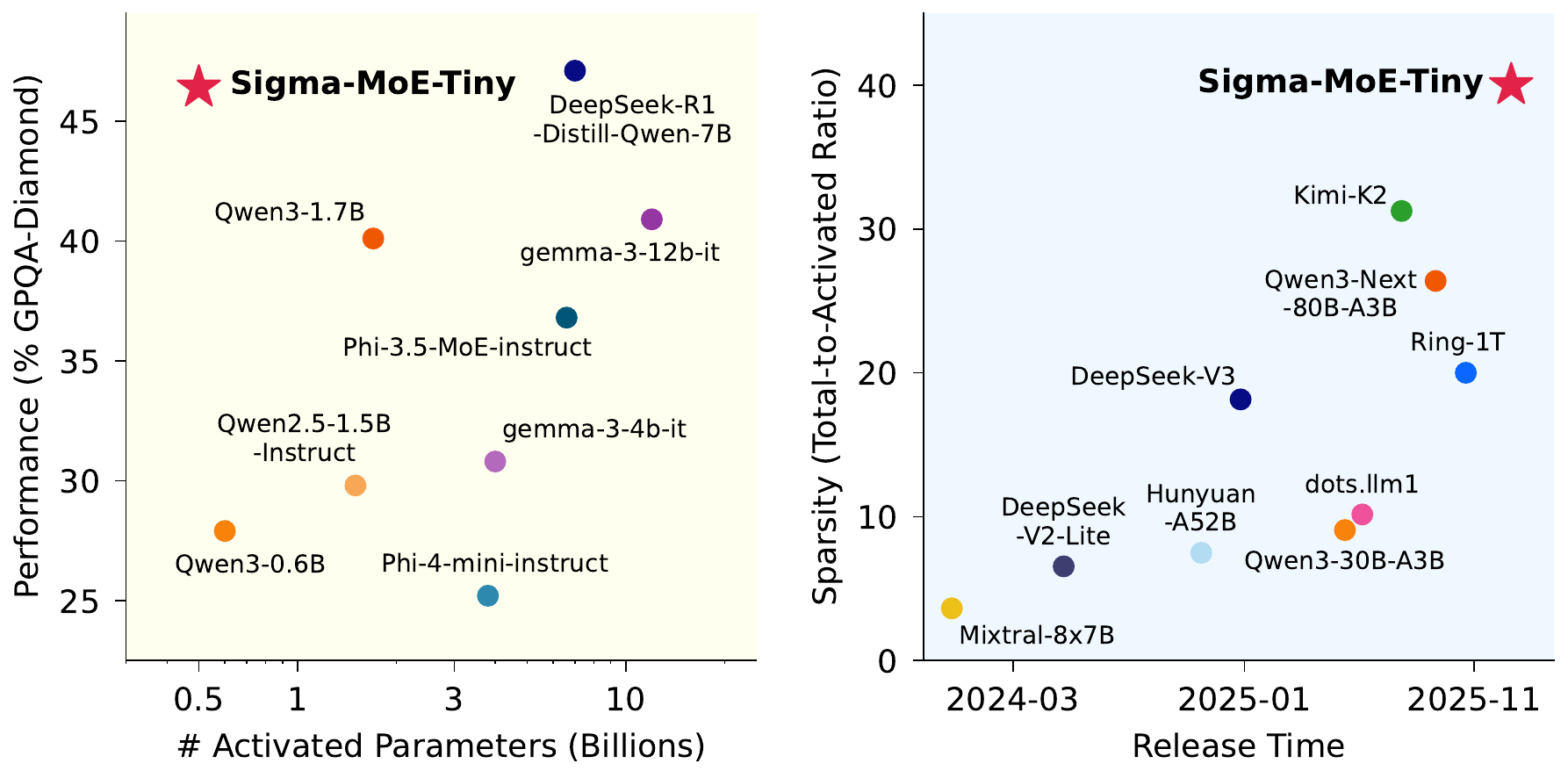}
    \caption{\textit{Left}: GPQA-Diamond accuracy vs. activated parameters across different open-source LLMs, demonstrating that \name{} achieves advanced capability with only 0.5B activated parameters. \textit{Right}: The trend of sparsity in mainstream MoE models over time is shown. Here, sparsity is defined as the ratio of total to activated parameters. With a total-to-activated ratio of 40:1, \name{} achieves the highest sparsity among existing open-source models.}
    \label{fig:overview}
\end{figure*}

%% file: section/1_intro.tex
\newpage
\section{Introduction}

The pursuit of Artificial General Intelligence (AGI) has long been a central aspiration in artificial intelligence research. Recent years have witnessed Large Language Models (LLMs) rapidly narrowing this gap. Proprietary frontier systems such as Gemini 3~\citep{gemini3}, GPT-5~\citep{gpt5}, and Claude 4~\citep{claude4}, together with leading open-source efforts including DeepSeek-V3~\citep{liu2024deepseek} and Qwen3~\citep{yang2025qwen3}, exemplify the remarkable pace of progress.
The ongoing advances in model and data scaling, combined with large-scale pre-training followed by high-quality supervised fine-tuning and reinforcement learning, have enabled them to develop emergent capabilities in complex understanding, generation, and reasoning.

Among current leading LLMs~\citep{liu2024deepseek, yang2025qwen3, huo2025dots, llama4}, the Mixture-of-Experts (MoE) architecture~\citep{fedus2022switch} has emerged as a defining trend for building efficient and powerful foundation models. By dynamically routing tokens to a small subset of experts, MoE models can achieve a vast parameter capacity while maintaining an economical computational cost, thereby enabling both scalability and efficiency. To further unlock the capabilities of MoE, the open-source community is actively developing increasingly sparse MoE models~\citep{huo2025dots, team2025kimi,yang2025qwen3,ling2025everystep}, demonstrating the potential of MoE as a fundamental paradigm for scaling next-generation LLMs.

In this work, we present \name{}, an MoE model with extremely high sparsity among existing open-source models, aiming to further push the limits of MoE sparsity. Following~\cite{dai2024deepseekmoe}, \name{} employs fine-grained expert segmentation, with up to 96 experts per MoE layer. To achieve super-high sparsity, only one expert is activated for each token. As a result, \name{} contains 20B parameters in total while activating merely 0.5B parameters per token, enabling highly efficient training and inference. Besides, we adopt Group Query Attention (GQA)~\citep{ainslie2023gqa} to reduce KV-cache overhead during inference, and combine it with QK-Norm~\citep{dehghani2023scaling} to ensure training stability.

A key challenge in training \name{} is maintaining expert load balance. Initially, we apply the auxiliary Load Balancing Loss (LBL)~\citep{qiu2025demons}. However, we observe that the widely-used LBL becomes ineffective in the lower layers under such extreme sparsity. Specifically, in this setting, the optimization of LBL tends to take a shortcut by pushing the gating probabilities of all experts toward a uniform distribution rather than balancing the token allocation fraction, thereby converging to an unintended minimum. To mitigate this issue, we propose a progressive sparsification schedule for \name{} training. During initial training, we activate more experts in the lower layers while maintaining the remaining layers at the target sparsity. In the later stages of training, we then switch all layers to the target sparsity. This approach not only effectively ensures expert load balance but also improves overall model performance.

During the pre-training phase, \name{} utilizes a high-quality and diverse corpus. The entire training process was highly stable, and we did not encounter any irrecoverable loss spikes. Regarding load balancing, all experts were maintained in relatively balanced utilization throughout the training process. In the post-training phase, we conduct supervised fine-tuning to extend \name{}’s context length and leverage long-CoT data to strengthen its reasoning capabilities. We adopt a multi-stage curriculum that progressively expands the model’s context window from 4K to 128K tokens and leverages training samples with increasing reasoning complexity at each stage. This curriculum-like progression enables the model to not only handle longer contexts but also develop stronger reasoning capabilities.

We evaluate both the pre-trained and post-trained versions of \name{} across a wide range of benchmarks.
Experimental results demonstrate that even with only 0.5B activated parameters, our pre-trained model still achieves top-tier performance among existing small-scale models. For the post-trained model, \name{} further delivers remarkable performance across diverse benchmarks, matching or even surpassing models several times larger in scale. As illustrated in Figure~\ref{fig:overview}, on GPQA-Diamond~\citep{gpqa}, \name{} attains leading performance comparable to dense models at the 7–10B scale. These results underscore the strong potential of extreme MoE sparsity for enhancing both model efficiency and overall capability. Moreover, we provide a comprehensive analysis and exploration of different load balancing strategies under extreme sparsity, offering deeper insights for building more effective sparse MoE architectures.

%% file: section/2_arch.tex
\section{Architecture}
The \name{} model adopts the widely-used decoder-only Transformer architecture~\citep{vaswani2017attention}, constructed by stacking $L$ layers of standard Transformer blocks. Each block consists of a self-attention module featuring with casual masks, followed by a Feed-Forward Network (FFN) module.
For attention, we adopt Group Query Attention (GQA)~\citep{ainslie2023gqa} to reduce the potentially enormous KV-cache overhead during the inference stage. Additionally, QK-Norm~\citep{dehghani2023scaling} is applied to the hidden states of both query and key prior to computing the attention map, which prevents the occurrence of excessively large attention logits during training. Regarding the FFN, we employ the popular Mixture-of-Experts (MoE) architecture. An MoE layer consists of a gating network and multiple experts, where each expert is identical to a two-layer FFN with a SwiGLU~\citep{shazeer2020glu} activation function. We use FP32 precision for computations in the gating network to ensure numerical stability, thereby enabling more accurate expert routing.
Following \citet{touvron2023llama}, we apply RMSNorm~\citep{zhang2019root} with pre-normalization to mitigate gradient vanishing issues during training.

\begin{wrapfigure}{r}{0.475\textwidth}
\centering
\vspace{-10pt}
\begin{tabular}{lr}
\toprule
\textbf{Configuration} & \textbf{Value} \\ \midrule
Hidden Size             & $1536$ \\
MoE Intermediate Size         & $768$ \\
\# Layers                  & $56$ \\
\# Heads (Q / KV)          & $16 / 4$ \\
\# Experts (Total / Activated)& $96 / 1$ \\
\# Params (Total / Activated) & $20\text{B} / 0.5\text{B}$ \\
\bottomrule
\end{tabular}
\captionof{table}{Model architecture of \name{}.}
\label{tab:arch}
\vspace{-5pt}
\end{wrapfigure}

\subsection{Super-high MoE Sparsity}
Early popular MoE-based LLMs often employ a limited number of experts (e.g., 8 or 16) to ensure better training stability. For instance, Mixtral-8x7B \citep{jiang2024mixtral} uses only 8 experts in total and activates 2 per token. However, this low-sparsity characteristic may lead to knowledge redundancy among experts and hinder their specialization, thereby preventing MoE models from reaching their upper-bound performance \citep{dai2024deepseekmoe}. Recently, many state-of-the-art MoE models, such as DeepSeek-V3~\citep{liu2024deepseek} and Qwen3~\citep{yang2025qwen3}, have demonstrated the effectiveness of using fine-grained expert segmentation~\citep{dai2024deepseekmoe}. In this context, the MoE layer adopts a larger number of smaller experts without increasing the overall parameter count. This improved sparsity further exploits the potential for expert specialization and enhances overall model performance. In this work, we further push the limit of MoE sparsity, aiming to achieve stronger capabilities with lower computational cost. Specifically, each MoE layer of \name{} contains up to \textbf{96 experts} in total, but with only \textbf{one expert} activated for each token, resulting in extremely high expert sparsity. The detailed model architecture of \name{} is provided in Table~\ref{tab:arch}. To further reduce the number of activated parameters, we employ the MoE architecture across all layers, without using standard dense FFNs in the lower~\citep{liu2024deepseek,huo2025dots} or intermediate~\citep{llama4} layers. By leveraging this super-high sparsity, our \name{} has a total of \textbf{20B parameters}, while only \textbf{0.5B parameters} are activated for each token. As shown in Figure~\ref{fig:overview}, it achieves a total-to-activated ratio of 40:1, which is the highest among existing open-source MoE models.

\subsection{Load Balance Considerations}
\label{sec:lbl}

\subsubsection{Expert Load Balancing}
We take the load balance of experts into consideration, as imbalanced loading will raise the risk of routing collapse~\citep{shazeer2017outrageously}. Moreover, since expert parallelism is typically employed during MoE training, imbalanced expert loading can also reduce computational efficiency. A typical solution is to introduce the Load Balancing Loss (LBL)~\citep{fedus2022switch} to mitigate these issues. The widely-used LBL considers the fraction of tokens $f_i$ routed to expert $E_i$ and the average gating probability $p_i$ of $E_i$, then the LBL is computed as the sum of the products of $f_i$ and $p_i$ over all $N_E$ experts, normalized by the number of experts:
\begin{equation}
\text{LBL} = N_E \sum_{i=1}^{N_E} f_i \cdot p_i.
\label{eq:lbl}
\end{equation}
Previous works mainly apply LBL at the sequence-level~\citep{liu2024deepseek} or micro-batch-level~\citep{fedus2022switch} (i.e., the statistical scope of $f_i$). Under such strict constraints, tokens from a specific domain may be routed uniformly across all experts, which can potentially inhibit expert specialization. To address this issue, following \citet{qiu2025demons}, we adopt a global-batch LBL to mitigate load imbalance. In this context, $f_i$ is synchronized across all parallel groups via an all-reduce operation to compute the average, resulting in a global-batch level statistic. By doing so, this LBL encourage expert load balance over the entire batch, thereby better promoting expert specialization.

\begin{figure*}[t]
    \centering
    \hspace{-0.015\textwidth}
    \begin{subfigure}[b]{0.8\textwidth}
        \centering
        \includegraphics[width=\textwidth]{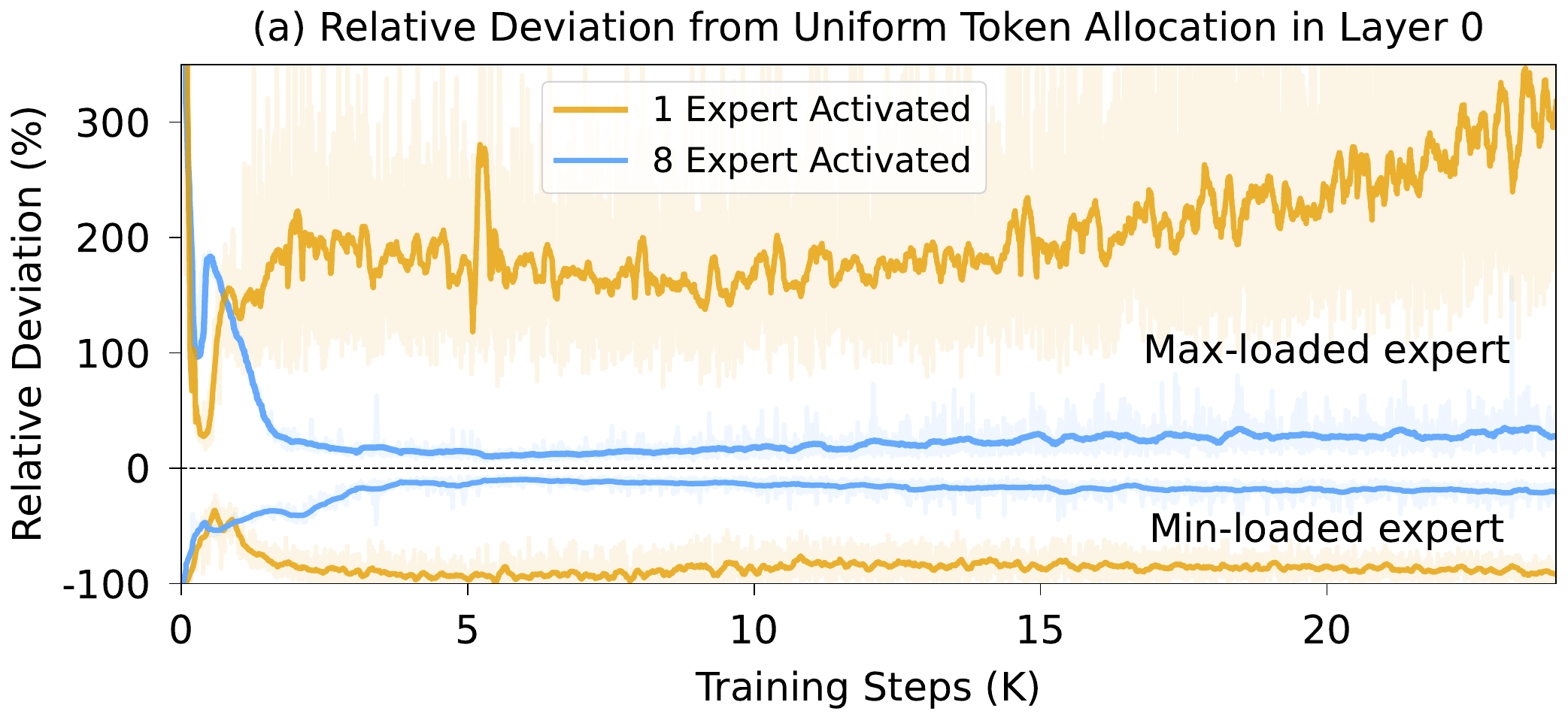}
        \label{fig:overview_top}
    \end{subfigure}
    
    \vspace{0.8pt}

    \hspace{-0.015\textwidth}
    \begin{subfigure}[b]{0.532\textwidth}
        \centering
        \includegraphics[width=\textwidth]{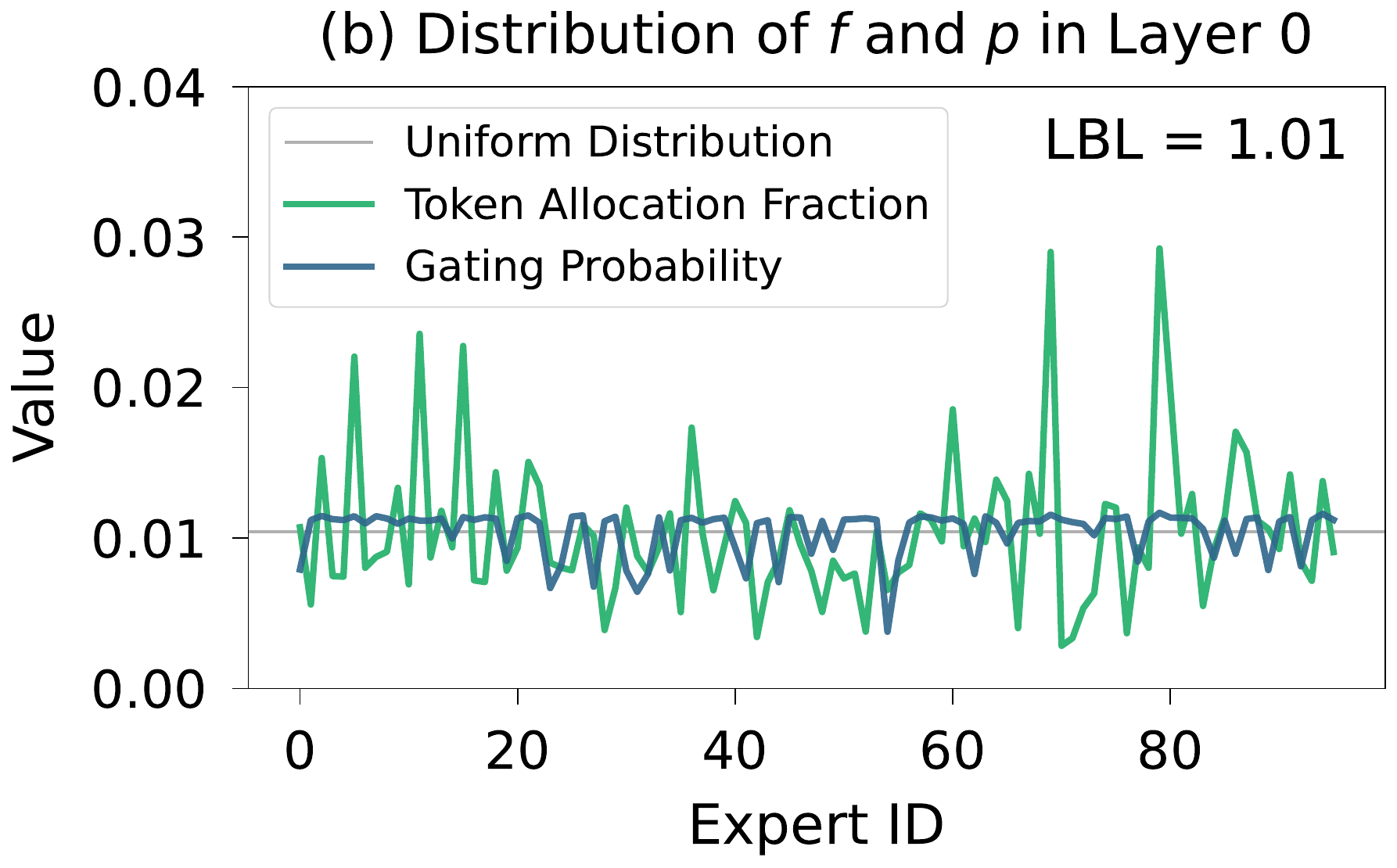}
    \end{subfigure}
    \hspace{0.015\textwidth}
    \begin{subfigure}[b]{0.437\textwidth}
        \centering
        \includegraphics[width=\textwidth]{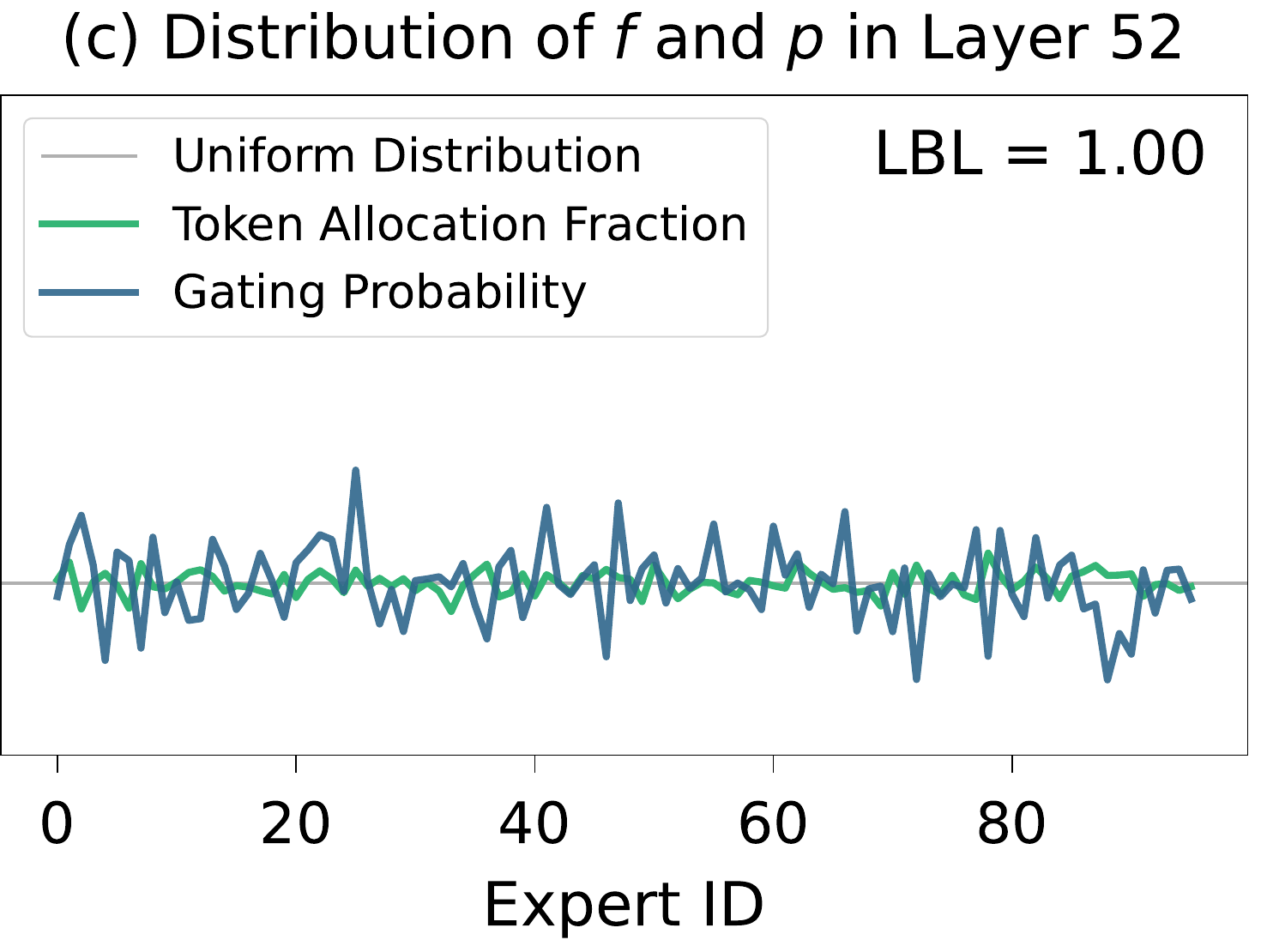}
    \end{subfigure}
    
    \caption{(a) Relative deviation from uniform token allocation is defined as the ratio between the difference of an expert's actual token count and the ideal uniform count, normalized by the uniform count. We report this deviation for the max-loaded and min-loaded experts in Layer 0. (b) and (c) show the distribution of token allocation fraction $f$ and gating probability $p$ across all experts in Layer 0 and Layer 52, respectively.}

    \label{fig:lbl}
\end{figure*}

\subsubsection{Progressive Sparsification Scheduling}
\label{sec:progressive}
In our initial experiments, we observe that simply applying the aforementioned LBL to our highly-sparse MoE architecture introduces a significant drawback: it leads to routing collapse in the lower layers. As shown in Figure~\ref{fig:lbl}(a), in the 0th layer, the min-loaded expert exhibits a relative deviation of nearly $-100\%$ from uniform token allocation, meaning it receives almost none of the tokens in each batch. In contrast, the max-loaded expert is routed close to $3\times$ the tokens expected under uniform allocation.

To understand this phenomenon, we track the token allocation fractions $f$ and the gating probabilities $p$ across all experts within a global batch during training. Our analysis reveals that under such extreme sparsity, the optimization objectives of the LBL differ between lower and higher layers. As shown in Figure~\ref{fig:lbl}(b) and (c), in the 52nd layer, the token allocation fractions $f$ are optimized as expected be approximately uniformly distributed across experts, while the gating probabilities $p$ remain non-uniform. In contrast, in the 0th layer, the gating probabilities $p$ are optimized toward uniformity, whereas the token allocation fractions $f$ remain highly non-uniform, contrary to the intended goal of balanced token allocation. Nevertheless, in both cases, the LBL approaches its ideal minimum.

We attribute this deficiency to the inherent characteristics of the LBL. In Equation~\ref{eq:lbl}, the desired goal of this loss is to optimize the token allocation fractions $f$ toward a uniform distribution. However, from the perspective of optimization, the LBL can reach its ideal minimum by making either $f$ or $p$ uniform. Under high sparsity, routing tokens in the lower layers becomes more difficult. Consequently, the LBL optimization takes a shortcut by driving $p$ towards a uniform distribution, resulting in an unintended minimum that fails to achieve true load balance.

To tackle this, we introduce a progressive sparsification schedule for \name{} training. The core idea is to start with a modest sparsity in lower layers when training from scratch and then transition to our proposed high sparsity later in the training process.
Specifically, during the early training phase, we activate more experts in the first 8 layers, while the remaining layers maintain the target sparsity (i.e., 1 expert out of 96). Considering that the impact of LBL ineffectiveness gradually diminishes in higher layers, the number of activated experts in the first 8 layers is set to [8, 8, 6, 6, 4, 4, 2, 2], thereby reducing the extra computational cost from increased activated parameters.

As shown in Figure~\ref{fig:lbl}(a), introducing modest sparsity into the lower layers substantially improves expert load balance throughout the training process. Furthermore, we find that this strategy can also preserve model performance when transitioning to the target sparsity ($\sim 25\%$ reduction in activated parameters, see discussion in Section~\ref{sec:analysis}), further demonstrating its effectiveness.

%% file: section/3_pre.tex
\section{Pre-training}

\subsection{Pre-Training Data}

\name{} is pre-trained on a diverse and high-quality dataset constructed from a mixture of public and proprietary sources. This dataset includes subsets of Nemotron-CC~\citep{su2024nemotron}, deduplicated DCLM~\citep{li2024datacomp}, and deduplicated FineWeb-Edu~\citep{penedo2024fineweb}, along with proprietary synthetic data. The dataset spans a wide range of domains, including general knowledge, mathematics, and coding, providing comprehensive coverage to enhance the model’s language understanding, knowledge retrieval, reasoning, and problem-solving capabilities.

\subsection{Training Hyper-Parameters}
We train \name{} using the AdamW optimizer~\citep{loshchilov2017decoupled}, with $\beta_{1}=0.9$, $\beta_{2}=0.95$ and $\epsilon=10^{-9}$. We use a weight decay of 0.1 and apply gradient clipping at 1.0. All learnable parameters are initialized from a normal distribution with a standard deviation of 0.02. We set the maximum sequence length to 4K during pre-training.
Following~\citet{liu2024deepseek}, we adopt a warmup–stable–decay learning rate schedule. The learning rate is linearly increased from 0 to $2.6\times10^{-4}$ during the first 2K steps. We keep the learning rate constant at $2.6\times10^{-4}$ in the first 60\% of the training corpus. Then, the learning rate decays to $1.6\times10^{-4}$ in the subsequent 30\% of the corpus, following a cosine decay schedule. Finally, we decay the learning rate to $2.6\times10^{-5}$ during the remaining 10\% of the corpus.
We also employ a batch size scheduling strategy, where the batch size is gradually increased from 1920 to 7680 during the first 40\% of the training corpus and then keeps 7680 in the remaining training. For expert load balancing, we set the coefficient of the global-batch LBL to $1\text{e-}3$. Regarding progressive sparsity scheduling, we apply a modest sparsity to the first 8 layers over the first 90\% of the training process, and then switch to the target sparsity in the remaining training.

\subsection{Infrastructure}

\paragraph{Hardware.}
\name{} is trained on NVIDIA A100-40GB GPUs. Each node contains 8 GPUs connected via NVSwitch, and nodes are interconnected through an InfiniBand fabric.

\paragraph{Training Stack.}
We leverage the training stack from~\citet{qu2025sigmaaiempoweredtrainingstack} to realize a reliable, stable and efficient training for \name{}.

\paragraph{Efficiency Optimization.}
The major challenge comes from \name{}'s extreme sparsity. Compared with existing models, \name{} has much smaller hidden sizes and MoE top-k values, which make underlying GPU kernels much less efficient given the same micro batch size and model parallelism configuration.
Our insight is that, when micro batch size is the same, the much smaller hidden sizes and MoE top-k values in \name{} significantly reduce per-GPU communication traffic required for MoE token routing in Expert Parallelism (EP), which is micro-batch-size * top-k * hidden-size each time. This allows the option of large micro batch size to improve kernel efficiency, with large EP scope to fit the model into limited 40GB single-GPU memory, and with still limited communication traffic inflation.
Accordingly, we set micro batch size to 8, with 4-way tensor parallelism and 96-way expert parallelism for \name{} pre-training.

\input{table/eval_base}

\subsection{Evaluation}

\subsubsection{Benchmarks}

To systematically assess the capabilities of the \name{} base model, we evaluate it on a broad collection of benchmarks covering . The benchmarks are organized as follows:

\begin{itemize}
  \item \textbf{General Tasks}: To assess world knowledge, we employ MMLU~\citep{mmlu}, MMLU-Pro~\citep{mmlupro}, and SuperGPQA~\citep{supergpqa}. To evaluate English reading comprehension and contextual reasoning, we adopt BigBenchHard (BBH)~\citep{bbh}, PIQA~\citep{piqa}, ARC~\citep{arc}, HellaSwag~\citep{hellaswag}, and WinoGrande~\citep{sakaguchi2021winogrande}. For assessing scientific knowledge, we use GPQA~\citep{rein2024gpqa}, which focuses on graduate-level scientific questions.

  \item \textbf{Mathematics Tasks}: We evaluate mathematical reasoning capabilities with GSM8K~\citep{gsm8k} for foundational arithmetic and MATH~\citep{math} for advanced problem solving.
  
  \item \textbf{Coding Tasks}: Code generation ability is assessed on HumanEval~\citep{humaneval} and MBPP~\citep{mbpp}.
\end{itemize}

We compare \name{}-Base with multiple base models, including the Qwen3~\citep{yang2025qwen3}, Gemma-3~\citep{team2025gemma}, and DeepSeek-V2~\cite{liu2024deepseekv2}. All models are evaluated using the same evaluation pipeline and the widely-used evaluation settings to ensure fair comparison.

\subsubsection{Results}

As shown in Table~\ref{tab:main}, even when activating only 0.5B parameters, \name{}-Base achieves strong performance across benchmarks compared to other counterparts with comparable or larger model scales. This demonstrates the effectiveness and efficiency brought by \name{}-Base's super-high MoE sparsity.  
Based on the overall evaluation results, we highlight several key conclusions:  
1) On general tasks, \name{}-Base clearly outperforms Qwen3-0.6B, Gemma-3-4B, and DeepSeek-V2-Lite. This shows that \name{}-Base has strong capabilities in world knowledge, reading comprehension, and contextual reasoning, reflecting the comprehensiveness and diversity of our pre-training data.
2) On mathematics and coding tasks, although we do not adopt a specialized stage for mathematical or code-specific training, \name{}-Base still demonstrates superior mathematical reasoning and code generation abilities compared to the baseline models. This result further highlights the enhanced specialization conferred by extreme MoE sparsity.

\subsection{Discussion}
\label{sec:analysis}

\subsubsection{Effect of Progressive Sparsification Scheduling}
The primary motivation of our progressive sparsification schedule is to mitigate expert load imbalance in lower layers. Beyond this, we also investigate its effect on overall model performance. As shown in Table~\ref{tab:token_training}, we compare two settings starting from the same intermediate checkpoint: one continuing with the initial sparsity and the other switched to the target sparsity. Notably, although converting to the target sparsity loses 0.15B activated parameters (approximately 25\%), the resulting performance is largely preserved. For example, at 200B continued training tokens, the baseline achieves 63.69\% accuracy on MMLU, while our approach still reaches 63.53\%. These results demonstrate that the proposed progressive sparsification schedule not only improves expert load balance during training but also preserves nearly the same performance with substantially fewer activated parameters, highlighting its effectiveness in enhancing training efficiency.

\input{table/ablation_progressive}

\subsubsection{Comparison of Different Load Balancing Strategies}

\begin{figure*}[h]
    \centering
    \hspace{-0.015\textwidth}
    \includegraphics[width=0.8\textwidth]{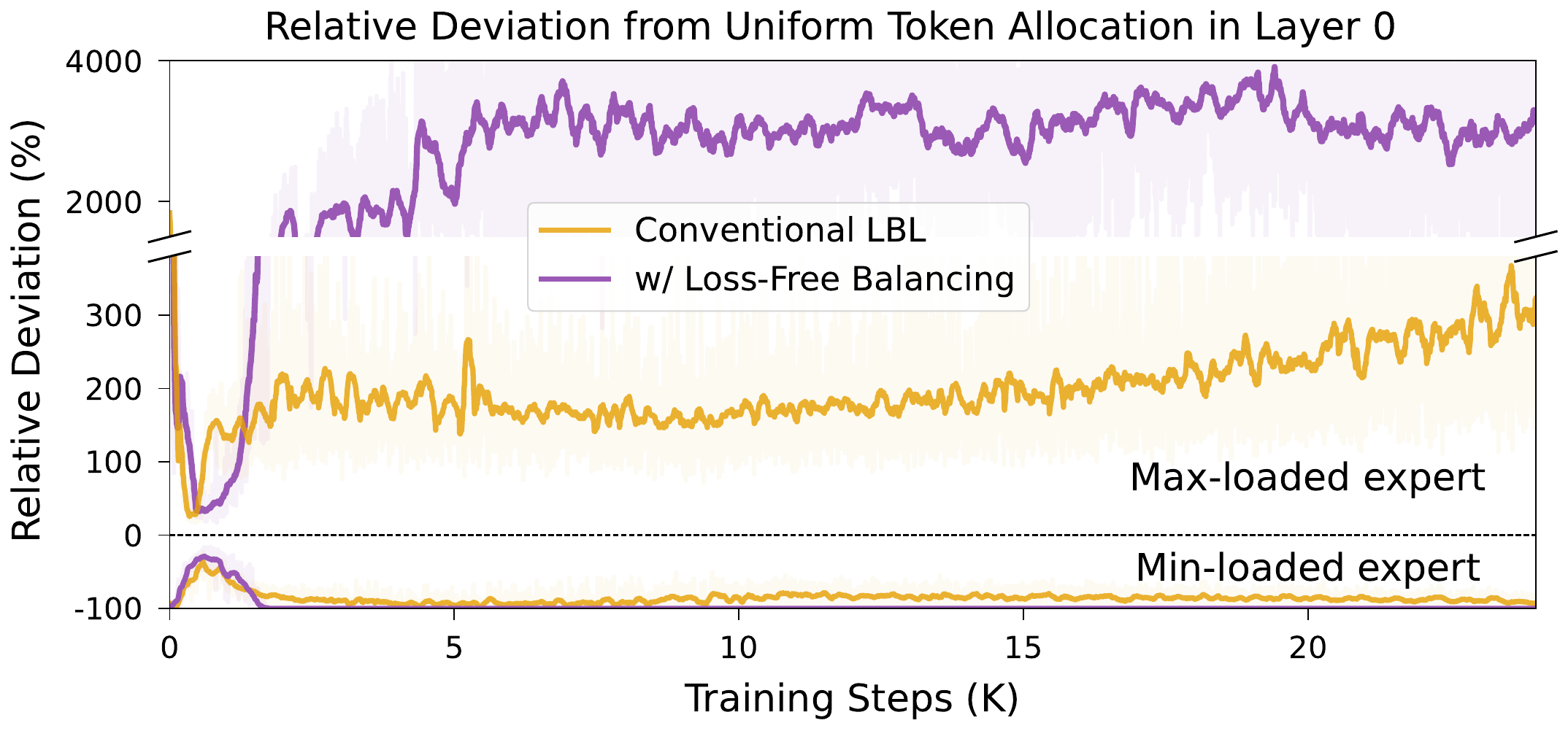}
    \caption{Relative deviation from uniform token allocation for the max-loaded and min-loaded experts in Layer 0. Introducing loss-free balancing strategy substantially aggravates load imbalance under the setting of 96 experts with 1 activated.}
    \label{fig:bias}
\end{figure*}

\citet{wang2024auxiliary} propose an auxiliary-loss-free approach to encourage load balance, which is also adopted in DeepSeek-V3~\citep{liu2024deepseek}. This method introduces an expert-wise bias to adjust the routing scores of each expert, where the bias is dynamically updated according to the recent load of the corresponding expert, thereby avoiding the introduction of interference gradients. 
In our preliminary experiments, we examine the effect of different load balancing strategies. Under super-high MoE sparsity, we observe that this loss-free approach can cause significant load imbalance in the lower layers. As shown in Figure~\ref{fig:bias}, compared with using only the conventional LBL (Equation~\ref{eq:lbl}), introducing this loss-free balancing strategy leads to the min-loaded expert consistently receiving zero tokens after 2K training steps, while the max-loaded expert is allocated nearly 40$\times$ the tokens expected under uniform allocation, accounting for almost half of all tokens in a global batch.

Further analysis reveals that the bias terms introduced by this strategy in the lower layers will continually increase as training progresses, eventually dominating the gating scores. As a result, the expert with the highest bias at each step receives the overwhelming majority of tokens. We attribute this deficiency to the following mechanism: according to this strategy, an expert's bias will be increased when it receives fewer tokens than the average. Considering that high MoE sparsity makes uniformly distributing tokens in the lower layers more difficult, this load imbalance basically persists throughout the training process and causes the bias terms to continue growing upward. Ultimately, the biases of all experts reach very high magnitudes. At this point, the portion of the gating scores provided by the router can no longer influence the final routing, leading to the expert with the highest bias to capture nearly all tokens.

\subsection{Exploring Native Load Balancing under High Sparsity}
As mentioned in Section~\ref{sec:lbl}, conventional LBL may become ineffective in lower layers under high sparsity. While our proposed progressive sparsification scheduling can address this issue, we also explore an alternative approach that achieves native load balancing without modifying the model architecture. To this end, we introduce a new LBL variant, called Top-1 LBL, which also follows the basic form in Equation~\ref{eq:lbl}. The core idea of this LBL is to directly optimize the L2 norm of the token allocation fraction across all experts, thereby theoretically avoiding the optimization bias present in conventional LBL. However, since the token allocation fraction is non-differentiable, we use the differentiable gating probabilities as an effective approximation, obtained by applying a temperature-scaled softmax to the routing logits. Formally, the Top-1 LBL is defined as:
\begin{equation}
\text{LBL}_{\text{Top-1}} = \frac{N_E \sum_{i=1}^{N_E} \hat{f}_i^2}{\bar{p}_\text{top-1}}, 
\label{eq:top1-lbl}
\end{equation}
where the token allocation fraction $\hat{f}_i$ for expert $i$ is computed as
\begin{equation}
\hat{f}_i = \frac{1}{N_B} \sum_{j=1}^{N_B} p_{i,j}, \quad 
p_{i,j} = \frac{\exp(\text{logits}_{i,j}/\tau)}{\sum_{k=1}^{N_E} \exp(\text{logits}_{k,j}/\tau)},
\end{equation}
and the average top-1 probability $\bar{p}_{\text{top-1}}$ in the denominator is defined as
\begin{equation}
\bar{p}_\text{top-1} = \frac{1}{N_B} \sum_{j=1}^{N_B} \text{Top-1}(p_{i,j}).
\end{equation}
Here, $N_B$ is the number of tokens in a batch, $\text{logits}_{i,j}$ is the routing logit of expert $i$ for token $j$, and $\tau$ is the softmax temperature. The role of the denominator $\bar{p}_\text{top-1}$ is to encourage maximizing the top-1 temperature-scaled gating probability for each token, thereby approximating the one-hot distribution obtained from top-1 sampling and thus providing an effective approximation of the token allocation fraction.

\definecolor{cinnamon}{RGB}{205,92,92}
\definecolor{bananayellow}{RGB}{255,225,53}

\pgfplotsset{
    /pgfplots/ybar legend/.style={
        /pgfplots/legend image code/.code={%
           \draw[##1,/tikz/.cd,yshift=-0.25em]
            (0cm,0cm) rectangle (7pt,0.8em);},
    },
}

\begin{wrapfigure}{r}{0.45\textwidth}
\centering
\pgfplotsset{width=6.5cm, height=4.5cm}
\begin{tikzpicture}  
    \begin{axis}[
        ybar,
        ymin=40, ymax=60,
        ytick={40, 45, 50, 55, 60},
        major x tick style = transparent,
        bar width=12pt,
        enlarge x limits=0.4,
        ylabel={Performance (\% MMLU)},
        xlabel={\# Training Tokens},
        symbolic x coords={0.6T, 0.8T, 1T},
        xtick=data,
        xticklabel style={yshift=4pt},
        y label style={at={(axis description cs:-0.11,0.5)},anchor=south},
        axis x line*=bottom,
        axis y line*=left,
        legend cell align=left,
        legend style={
            at={(0.7,0.7)},
            anchor=south east,
            column sep=1ex,
            font=\small,
            draw=none,
            legend columns=2,
            transpose legend,
        }
    ]
        \addplot[ybar, color=purple!80!black, fill=purple!20!white] coordinates {
            (0.6T, 52.05) (0.8T, 52.31) (1T, 53.00)
        };
        \addplot[ybar, color=teal!80!black, fill=teal!20!white] coordinates {
            (0.6T, 47.94) (0.8T, 48.82) (1T, 51.36)
        };
        \legend{
            Conventional LBL,
            Top-1 LBL
        }
    \end{axis}
\end{tikzpicture}
\vspace{-5pt}
\caption{Comparison of MMLU performance between Top-1 LBL and conventional LBL under different training token counts.}
\label{fig:top1-performance}
\vspace{-5pt}
\end{wrapfigure}

\begin{figure*}[t]
    \centering
    \hspace{-0.015\textwidth}
    \includegraphics[width=0.8\textwidth]{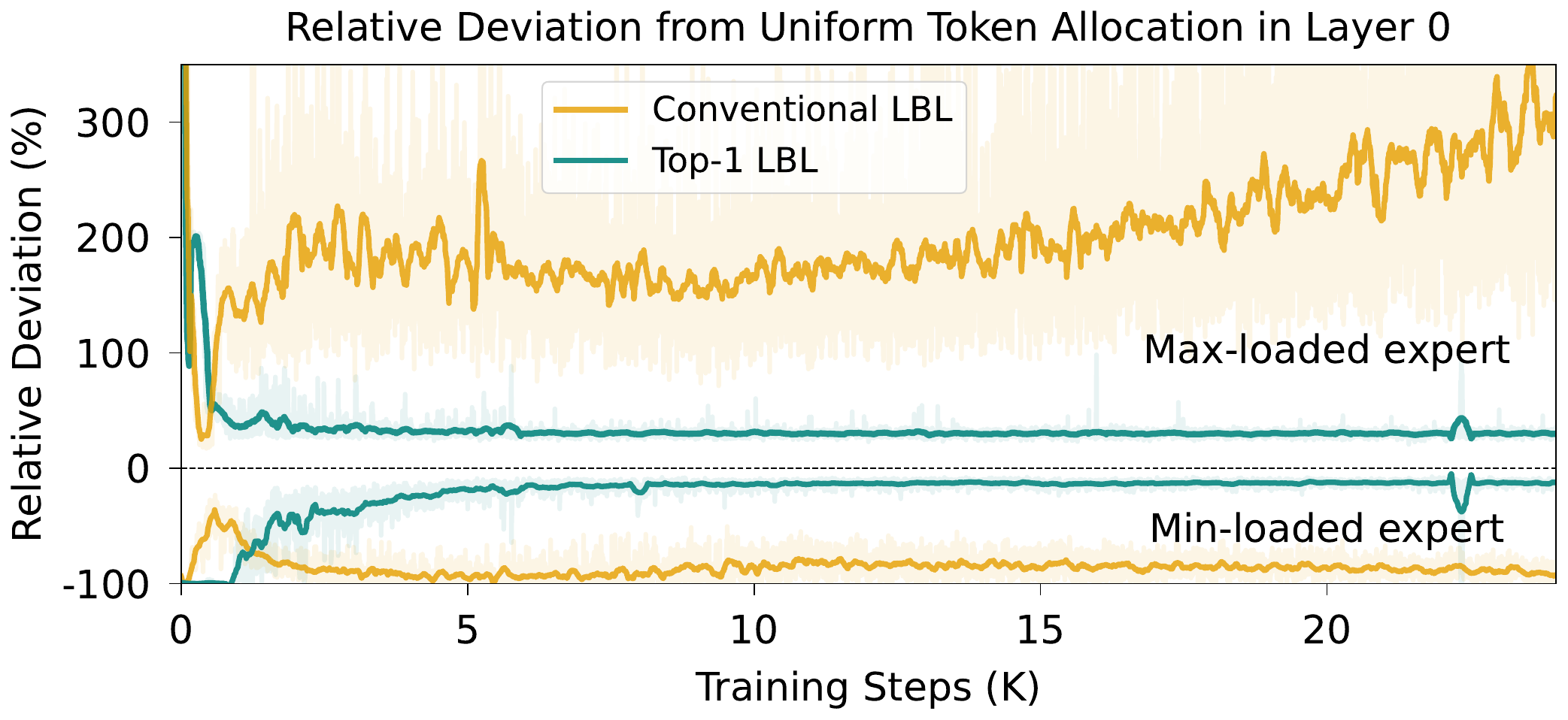}
    \caption{Relative deviation from uniform token allocation for the max-loaded and min-loaded experts in Layer 0. Compared to conventional LBL, introducing Top-1 LBL significantly improves load balancing under the setting of 96 experts with 1 activated.}
    \label{fig:top1}
\end{figure*}

We illustrate the effectiveness of Top-1 LBL on load balancing in Figure~\ref{fig:top1}. It can be seen that introducing this LBL significantly improves load balancing under high sparsity. Moreover, we also observe that it continues to balance expert utilization, steadily approaching a uniform token allocation. We further assess the benchmark performance of Top-1 LBL. As shown in Figure~\ref{fig:top1-performance}, we find that overly balanced expert utilization may sacrifice model performance. We attribute this issue to the inherent trade-off between load balance and model performance, as also pointed out in \cite{wang2024auxiliary}, where overly enforcing a balanced token allocation may introduce interfering gradients to language modeling training. For the challenge of better balancing this trade-off, we leave it as an important direction for future work.

%% file: table/eval_base.tex
\begin{table}[t]
    \centering
    \small
    \begin{tabular}{lc|ccc|c}
    \toprule
    \multirow{2}{*}{\centering \textbf{Benchmark {\tiny (Metric)}}} & \multirow{2}{*}{\textbf{\# Shots}} &   \textbf{Qwen3} & \textbf{Gemma-3} & \textbf{DeepSeek-V2} & \textbf{\name{}} \\
     & & \textbf{0.6B Base} & \textbf{4B Base} & \textbf{Lite} & \textbf{Base} \\
    \midrule
    Architecture & - & Dense & Dense & MoE & MoE \\
    \# Activated Params & - & 0.6B & 4B & 2.4B  & 0.5B \\
    \# Total Params & - & 0.6B & 4B & 15.7B & 20B \\
    \midrule
    \multicolumn{6}{c}{\textit{General Tasks}} \\
    \midrule
    MMLU {\tiny (EM)} & 5-shot & 52.81 & \underline{59.51} & 58.30 & \textbf{64.81} \\
    MMLU-Pro {\tiny (EM)} & 5-shot & 24.74 & \underline{29.23} & - & \textbf{38.13} \\
    BBH {\tiny (EM)} & 3-shot & 41.47 & \underline{51.70} & 44.10 & \textbf{63.23} \\
    PIQA {\tiny (Acc.)} & 0-shot & 69.86 & 80.09 & \underline{80.36} & \textbf{82.05} \\
    GPQA {\tiny (EM)} & 5-shot & \underline{26.77} & 24.24 & 24.55 & \textbf{27.68} \\

    ARC-C {\tiny (EM)} & 25-shot & 65.70 & \underline{74.66} & 71.76 & \textbf{80.29} \\
    HellaSwag {\tiny (EM)} & 10-shot & 53.57 & 77.76 & \textbf{80.55} & \underline{79.79} \\
    WinoGrande {\tiny (Acc.)} & 5-shot & 61.01 & \underline{72.53} & 71.11 & \textbf{76.09} \\
    \midrule
    \multicolumn{6}{c}{\textit{Mathematics Tasks}} \\
    \midrule
    GSM8K {\tiny (EM)} & 8-shot & \underline{59.59} & 43.97 & 41.10 & \textbf{71.65} \\
    MATH {\tiny (EM)} & 4-shot & \underline{32.44} & 26.10 & 17.10 & \textbf{36.88} \\

    \midrule
    \multicolumn{6}{c}{\textit{Coding Tasks}} \\
    \midrule
    HumanEval {\tiny (Pass@1)} & 0-shot & 29.27 & \underline{35.98} & 29.90 & \textbf{42.07} \\
    MBPP {\tiny (Pass@1)} & 3-shot & 36.60 & \underline{46.40} & 43.20 & \textbf{47.00} \\
    \bottomrule
    \end{tabular}
    \caption{
        Performance comparison among \name{}-Base and other base models across multiple domains. The highest and second-best scores are shown in bold and underlined, respectively.}
    \label{tab:main}
\end{table}

%% file: table/ablation_progressive.tex
\begin{table}[h]
    \centering
    \small
    \begin{tabular}{ccccccc}
        \toprule
         \multirow{2}{*}{\textbf{Setting}} & \multirow{2}{*}{\makecell{\textbf{\# Activated} \\ \vspace{-2pt} \textbf{Params}}}
 & \multicolumn{5}{c}{\textbf{\# Continued Training Tokens}} \\
        \cmidrule(lr){3-7}
          &  &  40B & 70B & 100B & 130B & 200B \\
        \midrule
        Maintain Initial Sparsity & 0.65B & 63.47 & 63.57 & 63.59 & 63.54 & 63.69 \\
        Convert to Target Sparsity & 0.50B & 62.99 & 63.06 & 62.35 & 63.04 & 63.53 \\
        \bottomrule
    \end{tabular}
    \caption{MMLU performance comparison between maintaining initial sparsity and converting to target sparsity. Both settings start from the same intermediate checkpoint for continued training.}

    \label{tab:token_training}
\end{table}

%% file: section/4_post.tex
\section{Post-training}

We perform post-training on \name{}-Base, during which we extend the context length and leverage Long-CoT data to strengthen its reasoning ability. This process aims to examine how ultra-sparse MoE architectures perform on practical downstream tasks.

\input{table/posttrain_para}
\subsection{Progressive Long-Context Extension} 
The \name{}-Base supports a maximum context length of 4K tokens, which we progressively extend during post-training. Specifically, we collected long-CoT datasets with sequence lengths below 16K, 32K, and 128K in three progressive stages, gradually expanding the model’s context window from 4K to 128K. 

To further improve the model’s long-context reasoning capability, we increased the RoPE base frequency from 10,000 to 1,000,000. Moreover, for samples significantly shorter than the target context lengths (16K, 32K, or 128K), we concatenated them when appropriate to maximize the utilization of the available context window and enhance training efficiency. 

\subsection{Supervised Fine-Tuning}
\input{table/think_mode} 

During the supervised fine-tuning stage, we progressively increase the difficulty of training samples in conjunction with the extension of context length. 
In particular, datasets with longer contexts (e.g., 128K) typically include more complex and reasoning-intensive problems to better exploit the model’s extended context capacity, while the 16K dataset primarily consists of questions solvable through relatively simple reasoning. 
This curriculum-like progression enables the model to not only handle longer contexts but also develop stronger reasoning capabilities for complex tasks. 

Building on this curriculum-like design, we implement the supervised fine-tuning through four stages. Specifically, in the first stage, the model is trained on 16K-length data that includes both Long-CoT and Short-CoT samples. As shown in \ref{tab:think_mode}, these two types of data differ in their construction: for Long-CoT samples, we introduced an additional \texttt{thinking prompt} in the system prompt to encourage explicit reasoning traces, whereas for Short-CoT samples, we left the thinking content empty and used the placeholder \texttt{<think></think>} to indicate the absence of an extended reasoning process.

The second and third stages employ datasets with context lengths of 32K and 128K, respectively, predominantly consisting of Long-CoT samples. In the final stage, we fine-tune the model on high-quality and diverse subset of 32K-length Long-CoT data, further consolidating its performance within the 32K context window. 

Across all stages, we maintained a balanced composition of data domains, with the ratio of mathematics : code : science : others set to 3.5 : 3.5 : 2 : 1, ensuring both domain diversity and representational consistency during supervised fine-tuning. The detailed training configurations for each stage are summarized in Table~\ref{tab:post-training-para}.

\subsection{Evaluation}
\input{table/posttrain_result}

\paragraph{Evaluation Tasks.} To comprehensively evaluate the overall capabilities of the post-trained models, we assess their performance across complex reasoning, code understanding and reasoning, and multi-domain knowledge reasoning using widely adopted evaluation tasks. The tasks are organized as follows:

\begin{itemize}
\item \textbf{General Tasks}: To evaluate broad knowledge and reasoning ability across diverse subjects, we employ MMLU-Redux~\citep{mmluredux2024we}, MMLU-Pro~\citep{wang2024mmlupro}, and GPQA-Diamond~\citep{gpqa}. Since GPQA-Diamond contains a relatively small number of questions, we report the average performance over 8 independent runs (\emph{avg@8}) to ensure evaluation stability.

\item \textbf{Mathematics Tasks}: To measure mathematical reasoning ability, we adopt Math-500~\citep{math2021}, AIME24, and AIME25~\citep{aime}. Both AIME24 and AIME25 consist of 30 competition-style problems, and we compute the mean accuracy over 16 independent runs (\emph{avg@16}) as the final result.

\item \textbf{Coding Tasks}: To assess programming and problem-solving competence, we use HumanEval~\citep{chen2021humaneval} and LiveCodeBench~\citep{jain2024livecodebench} ($\leq$ 202505).
\end{itemize}

\paragraph{Baselines.} 
For \name{}, we compare against DeepSeek-R1-Distill-Qwen-7B~\citep{guo2025deepseek}, DeepSeek-R1-Distill-Llama-8B~\citep{guo2025deepseek}, Phi-3.5-MoE~\citep{abdin2024phi3}, and Qwen3-1.7B~\citep{yang2025qwen3}. This set of baselines covers both dense and mixture-of-experts architectures of comparable or slightly larger parameter scales, allowing for a fair and comprehensive evaluation of \name{}’s efficiency-performance trade-off.

\paragraph{Hyperparameters Settings.} For \name{}, we perform inference following the format shown in Table~\ref{tab:think_mode}.
We adopt sampling hyperparameters with temperature = 0.6, top-p = 0.95, top-k = 20, and set the max position embeddings to 131,072.
Each response consists of two parts: a reasoning (think) section and a final summary answer section.
During inference, we allocate a thinking budget of 32,768 tokens for the reasoning section.
Once the model’s generated reasoning exceeds this limit, we append a closing phrase along with the special token \texttt{</think>}, prompting the model to directly produce its final answer based on the current reasoning content.
An additional 4,096-token window is then reserved for generating the final summary answer.

\paragraph{Summary of Evaluation Results.}

Table~\ref{tab:sigma-tiny-posttrain} present the evaluation results of \name{} against various popular models across multiple benchmarks. 
Despite contains only 0.5B active parameters, \name{} demonstrates outstanding performance across diverse benchmarks. 
In general domains, it exhibits a strong breadth of knowledge, achieving performance comparable to the 7B-parameter DeepSeek-R1-Distill-Qwen. 
On mathematical reasoning tasks, \name{} achieves 94.6\% on the Math-500 benchmark and attains 65.4\% and 48.8\% accuracy on AIME24 and AIME25, respectively, surpassing baseline models such as Qwen3-1.7B and Phi-3.5-MoE.
On the coding tasks, \name{} exhibits competitive capability, reaching 42.2\% on LiveCodeBench, on par with the 8B-parameter DeepSeek-R1-Distill-Llama.
Overall, these results suggest that super-high MoE sparsity, once properly post-trained, can match or even surpass the performance of significantly larger dense and MoE counterparts, while maintaining excellent generalization and reasoning efficiency. This underscores the promising potential of super-high sparse MoE architectures.

%% file: table/posttrain_para.tex
\begin{table}[!h]
\small
\centering
\setlength{\tabcolsep}{7mm}
\begin{tabular}{c|ccccc}
\toprule
& \textbf{Stage \uppercase\expandafter{\romannumeral 1}} & \textbf{Stage \uppercase\expandafter{\romannumeral 2}} & \textbf{Stag \uppercase\expandafter{\romannumeral 3}} & \textbf{Stage \uppercase\expandafter{\romannumeral 4}} \\
\midrule
Sequence Length & 16,384 & 32,768 & 131,072 & 32,768 \\
Batch Size & 128 & 96 & 64 & 64  \\
Max LR & 2.6e-5 & 1.5e-5 & 5.0e-6 & 1.0e-6 \\
Min LR & 1.5e-5 & 5.0e-6 & 5.0e-7 & 1.0e-7 \\
\bottomrule
\end{tabular}
\caption{Training Recipe for Post-training Alignment.}
\label{tab:post-training-para}
\end{table}

%% file: table/think_mode.tex
\begin{table}[tbp]
\centering
\adjustbox{center=\textwidth}{
\small
\begin{tabular}{@{}c|c@{}}
\toprule
\bf Long-CoT Data & \bf Short-CoT Data \\
\midrule
\tabincell{l}{\texttt{<|system|>system prompt}\\\texttt{thinking prompt<|end|>}\\\texttt{<|user|>user prompt<|end|>}\\
\texttt{<|assistant|>}\\
\texttt{<think>}\\
\texttt{thinking content}\\
\texttt{</think>}\\
\texttt{assistant response<|end|>}\\
}
& 
\tabincell{l}{\texttt{<|system|>system prompt<|end|>}\\\texttt{<|user|>user prompt<|end|>}\\
\texttt{<|assistant|>}\\
\texttt{<think>}\\
\texttt{</think>}\\
\texttt{assistant response<|end|>}\\
}
\\
\toprule
\bf Inference w/ Thinking & \bf Inference w/o Thinking \\
\midrule
\tabincell{l}{\texttt{<|system|>system prompt} \\\texttt{thinking prompt<|end|>}\\\texttt{<|user|>user prompt<|end|>}\\
\texttt{<|assistant|>}\\
\texttt{<think>}\\
\texttt{\{thinking budget\}}
}
& 
\tabincell{l}{\texttt{<|system|>system prompt<|end|>}\\\texttt{<|user|>user prompt<|end|>}\\
\texttt{<|assistant|>}\\
\texttt{<think>}\\
\texttt{</think>}\\
}
\\
\bottomrule
\end{tabular}
}
\caption{Illustration of the data construction and inference formats. \textit{Top:} The construction formats of \textbf{Long-CoT} and \textbf{Short-CoT} data. Long-CoT samples include an additional \textbf{thinking prompt} in the system prompt to elicit explicit reasoning traces, while Short-CoT samples leave the thinking content empty, denoted as \texttt{<think></think>}. \textit{Bottom:} The prompt formats used during inference in the \textbf{with(w/)} and \textbf{without(w/o) think} modes. The with-think mode introduces a \textbf{thinking budget} that constrains the reasoning length before producing the final answer.}
\label{tab:think_mode} 
\end{table}

%% file: table/posttrain_result.tex
\begin{table}[]
\centering
\setlength{\tabcolsep}{3pt}
\resizebox{\textwidth}{!}{
\scriptsize
\begin{tabular}{@{}lccccc@{}}
\toprule
 {\centering \textbf{Benchmark {\tiny (Metric)}}} & \tabincell{c}{\textbf{DeepSeek-R1}\\\textbf{-Distill-Qwen-7B}} 
 & \tabincell{c}{\textbf{DeepSeek-R1}\\\textbf{-Distill-Llama-8B}} 
 & \textbf{Qwen3-1.7B} 
 & \textbf{Phi-3.5-MoE}
 & \tabincell{c}{\textbf{Sigma-MoE}\\\textbf{-Tiny}} \\
\midrule
Architecture & Dense & Dense & Dense & MoE & MoE \\
\# Activated Params & 7B & 8B & 1.7B & 6.6B & 0.5B \\
\# Total Params & 7B & 8B & 1.7B & 42B & 20B \\
\midrule
\multicolumn{6}{c}{\textit{General Tasks}} \\
\midrule
\addlinespace[2pt]
MMLU-Redux\,{\tiny (avg@1)} & 68.5 & 66.4 & 73.9 & \underline{78.6} & \textbf{79.8} \\
MMLU-Pro\,{\tiny (avg@1)} & 52.0 & 52.7 & \underline{58.6} & 54.3 & \textbf{63.7} \\
GPQA-Diamond\,{\tiny (avg@8)} & \textbf{47.1} & 43.2 & 40.1 & 36.8 & \underline{46.4} \\
\midrule
\multicolumn{6}{c}{\textit{Mathematics Tasks}} \\
\midrule
\addlinespace[2pt]
MATH-500\,{\tiny (avg@1)} & 92.8 & 89.1 & \underline{93.4} & 59.5 & \textbf{94.6} \\
AIME'24\,{\tiny (avg@16)} & \underline{55.5} & 50.4 & 48.3 & 13.3 & \textbf{65.4} \\
AIME'25\,{\tiny (avg@16)} & \underline{39.2} & 27.8 & 36.8 & 6.7 & \textbf{48.8} \\
\midrule
\multicolumn{6}{c}{\textit{Coding Tasks}} \\
\midrule
\addlinespace[2pt]
HumanEval\,{\tiny (avg@1)} & 64.0 & 73.2 & 70.1 & \underline{75.0} & \textbf{79.9} \\
LiveCodeBench v6\,{\tiny (avg@1)} & 35.7 & \textbf{42.5} & 33.2 & 10.5 & \underline{42.2} \\
\bottomrule
\end{tabular}
}
\caption{Performance comparison of \name{} and baseline models across multiple domains. The highest and second-best scores are shown in bold and underlined, respectively.}
\label{tab:sigma-tiny-posttrain}
\end{table}

%% file: section/5_con.tex
\section{Conclusion}

In this work, we introduce \name{}, an MoE model that achieves the highest sparsity among existing open-source models. By activating only one expert per token out of 96 experts in each MoE layer, \name{} reaches a total of 20B parameters while requiring only 0.5B parameters activated per token. To address the inherent load imbalance in highly sparse MoE architectures, we identify the limitations of conventional load balancing loss under extreme sparsity and propose a progressive sparsification schedule that stabilizes training and ensures balanced expert utilization. Leveraging a high-quality pre-training corpus and a multi-stage post-training curriculum, \name{} demonstrates robust language understanding and reasoning capabilities.
Comprehensive evaluations show that \name{} achieves leading performance among small-scale models and remains competitive with much larger systems across a diverse set of benchmarks. These results underscore the promise of pursuing extreme MoE sparsity as a new scaling direction for next-generation LLMs, offering a practical path toward efficient yet capable foundation models.

%% file: section/appendix.tex
\section{Author List}

\paragraph{Core Contributors}\mbox{}

Qingguo Hu*, Zhenghao Lin, Ziyue Yang, Yucheng Ding*, Xiao Liu, Yuting Jiang, Ruizhe Wang*, Tianyu Chen, Zhongxin Guo, Yifan Xiong, Rui Gao, Lei Qu, Jinsong Su*, Peng Cheng, Yeyun Gong

\paragraph{Acknowledgments}\mbox{}

Anna Daly, Boris Pinzur, Guoshuai Zhao, Haoran Deng*, Han Zhang*, Hao Ni*, Hongyi He*, Hossein Pourreza, Jian Jiao, Joe Chau, Julia Katilevsky, Lianhai Ren*, Logan Cope, Luis Martinez Castillo, Qinzheng Sun*, Ray Jui-Hao Chiang, Ryn Vinogradova, Shuai Lu*, Xiao Liang*, Xingjian Zhang*, Yaoxiang Wang*, Yasen Hu*, Yelong Shen, Ying Xin, Yu Yan*, Zijie Chen*

\vspace{10pt}

An asterisk (*) next to a name indicates individuals who are interns, vendors, or those who have left the team. Within the acknowledgments, authors are listed alphabetically by first name.